\documentclass{article}

\usepackage{PRIMEarxiv}

\usepackage[utf8]{inputenc} % allow utf-8 input
\usepackage[T1]{fontenc}    % use 8-bit T1 fonts
\usepackage{hyperref}       % hyperlinks
\usepackage{url}            % simple URL typesetting
\usepackage{booktabs}       % professional-quality tables
\usepackage{amsfonts}       % blackboard math symbols
\usepackage{nicefrac}       % compact symbols for 1/2, etc.
\usepackage{microtype}

\usepackage{lipsum}
\usepackage{fancyhdr}       % header
\usepackage{graphicx}       % graphics
\graphicspath{{media/}}     % organize your images and other figures under media/ folder
\usepackage{setspace}
\usepackage{subcaption}
\usepackage{amsmath}
\usepackage{algpseudocode}
\usepackage{algorithm}
\usepackage{booktabs}
\usepackage{array}
\usepackage[utf8]{inputenc}
\usepackage{tcolorbox}
\usepackage{mdframed}
\definecolor{mycolor}{RGB}{210, 235, 255} % You can adjust these RGB values to change the color
\definecolor{mycolor2}{RGB}{255, 224, 178}
\usepackage{xcolor}
\usepackage{setspace}
\usepackage{multicol}

\usepackage{subcaption}

\DeclareCaptionFormat{mysub}{#3}  % Only keep the caption text, discard the rest
\usepackage{titlesec}
\titlespacing*{\section}{0pt}{0.3\baselineskip}{0.2\baselineskip}
\usepackage{graphicx}
\usepackage{tikz}
\tikzset{every node/.style={inner sep=0pt, outer sep=0pt}} 

\title{EMA-SAM: Exponential Moving-average for SAM-based PTMC Segmentation}

\author{
Maryam Dialameh\textsuperscript{1}, Hossein Rajabzadeh\textsuperscript{1}, Jung Suk Sim\textsuperscript{2,3}, and Hyock Ju Kwon \textsuperscript{1}\\
\textsuperscript{1}\textit{Department of Mechanical and Mechatronics Engineering, University of Waterloo, Canada} \\
\textsuperscript{2}\textit{Department of Radiology, Withsim Clinic, Seongnam, and \textsuperscript{3} Ewha Womans University Medical Center, Seoul, Korea} \\
\texttt{\{maryam.dialameh, hossein.rajabzadeh, hjkwon\}@uwaterloo.ca}, \texttt{jungsuk.sim@gmail.com}
}

\begin{document}
\maketitle
\begin{abstract}
Papillary thyroid microcarcinoma (PTMC) is increasingly managed with radio-frequency ablation (RFA),
yet accurate lesion segmentation in ultrasound videos remains difficult due to low contrast,
probe-induced motion, and heat-related artifacts. The recent Segment Anything Model 2 (SAM-2)
generalizes well to static images, but its frame-independent design yields unstable predictions and
temporal drift in interventional ultrasound. We introduce \textbf{EMA-SAM}, a lightweight extension
of SAM-2 that incorporates a confidence-weighted exponential moving average pointer into the memory
bank, providing a stable latent prototype of the tumour across frames. This design preserves
temporal coherence through probe pressure and bubble occlusion while rapidly adapting once clear
evidence reappears. On our curated PTMC-RFA dataset (124 minutes, 13 patients), EMA-SAM improves
\emph{maxDice} from 0.82 (SAM-2) to 0.86 and \emph{maxIoU} from 0.72 to 0.76, while reducing false
positives by 29\%. On external benchmarks, including VTUS and colonoscopy video polyp datasets,
EMA-SAM achieves consistent gains of 2--5 Dice points over SAM-2. Importantly, the EMA pointer adds
\textless0.1\% FLOPs, preserving real-time throughput of $\sim$30\,FPS on a single A100 GPU. These
results establish EMA-SAM as a robust and efficient framework for stable tumour tracking, bridging
the gap between foundation models and the stringent demands of interventional ultrasound. Codes are available here \hyperref[code]{https://github.com/mdialameh/EMA-SAM}.
\end{abstract}

% % keywords can be removed
% \keywords{First keyword \and Second keyword \and More}

\section{Introduction}
\label{intro}
Papillary thyroid micro-carcinoma (PTMC)—malignant nodules less than or equal to $1.0 \mathrm{cm}$—may appear clinically indolent, yet inadequate treatment can still lead to recurrence or distant spread \cite{dideban2016thyroid,kaliszewski2020risk}. Radio-frequency ablation (RFA) has emerged as a minimally invasive alternative to open surgery, thermally ablating the tumor under real-time ultrasound guidance \cite{bernardi2021current,orloff2022radiofrequency}. In practice, however, two factors undermine the precision of this technique: (i) the operator must repeatedly relocate a low-contrast lesion and delineate its margins despite probe motion and tissue compression, and (ii) heat-induced micro-bubbles generate acoustic shadows that obscure the residual tumor during subsequent sonications \cite{kaliszewski2019papillary}. These difficulties are compounded by the intrinsic resolution limits of B-mode ultrasound and its heavy dependence on operator skill \cite{mohammadkarim2018hemodynamic,afrin2023deep}. Because bubbles can persist for several minutes, clinicians risk leaving microscopic foci untreated or delivering additional ablations that increase collateral thermal injury \cite{huang2020phase,shi2019inflammation}. An automated system capable of segmenting PTMC robustly and in real time throughout the entire RFA video stream would therefore address a critical unmet need, providing consistent lesion localization, guiding energy deposition, and ultimately improving oncological and safety outcomes

The Segment Anything Model (SAM) \cite{kirillov2023segment} introduced a promptable image–only architecture that couples a powerful masked autoencoder backbone with a lightweight mask decoder. Trained on billions of synthetic and manual masks, SAM generalises remarkably well across domains, enabling zero-shot delineation of previously unseen objects from a single point, box, or coarse mask. While this flexibility has sparked rapid adoption in medical imaging, SAM’s frame-independent design provides no mechanism to link ambiguous predictions over time, forcing any video application to treat each frame in isolation. SAM-2 \cite{ravi2024sam} remedies this limitation by embedding the same promptable decoder inside a memory-augmented transformer that shares information across frames. A FIFO memory bank stores both spatial feature maps and compact “object-pointer” tokens, while a dedicated presence head allows the model to output “no object” whenever the target is occluded. Crucially, SAM-2 conditions each new frame on these memories \emph{before} decoding, yielding temporally coherent masks without external tracking. This principled integration of short- and long-term context is the key advantage over image-only SAM and is exactly what is needed to maintain a stable PTMC outline through probe movement, tissue compression, and bubble-induced occlusion in RFA videos.

Although SAM-2’s memory bank delivers frame-to-frame coherence, its \emph{greedy}, first-in-first-out updates can corrupt the very context it relies on whenever the PTMC becomes invisible—a frequent occurrence during RFA as probe pressure displaces tissue or heat generates micro-bubble shadows. In such moments the network still overwrites its memory with noisy, low-confidence tokens, so that when the lesion re-enters view the decoder is guided by a distorted prior, producing flickering masks or complete tracking failure. To overcome this weakness we introduce \textbf{EMA-SAM}—an \emph{E}xponential \emph{M}oving-average \emph{A}uxiliary pointer that lives alongside SAM-2’s object tokens and is updated each frame through a confidence-weighted momentum rule. The pointer is never dropped from the FIFO queue and is softly over-weighted during cross-attention, acting as a stable latent prototype of the tumor. Consequently, EMA-SAM preserves a reliable PTMC representation through seconds-long disappearances and bubble occlusions, yet immediately adapts once high-confidence imagery returns, yielding markedly smoother and more accurate segmentation across the full RFA sequence.

\section{Related Work}

Segment Anything Model (SAM) by Kirillov et al. \cite{kirillov2023segment} marked a major step toward universal segmentation by introducing a promptable vision foundation model trained on over one billion masks. SAM’s architecture, comprising a prompt encoder, image encoder, and mask decoder, enables segmentation of arbitrary objects in static images based on minimal user input, such as points, boxes, or masks. To extend SAM for accounting temporal information, SAM‑2 is proposed, extending SAM to the video domain by introducing a memory-augmented framework that incorporates past segmentation information through a memory bank. Ding et al. \cite{ding2024sam2long} introduce SAM2Long, which enhances SAM‑2 with a training-free memory tree to mitigate greedy error accumulation over long video sequences. Instead of relying on a single prediction per frame, SAM2Long maintains multiple segmentation pathways and performs a constrained tree search to select the highest-confidence sequence. The Medical SAM Adapter \cite{wu2025medical} inserts tiny adapter blocks into SAM, fine-tuning only ~2\% of parameters.
It reshapes 2-D embeddings for volumetric data and adds prompt-conditioned modulation, boosting performance across multi-modal CT/MRI tasks without full retraining. \textbf{SAM2-UNet} \cite{xiong2024sam2} repurposes the Hiera backbone of SAM-2 as a frozen encoder and appends a lightweight U-shaped decoder with adapter layers, enabling parameter-efficient fine-tuning.
This hybrid retains SAM-2’s rich pre-training while matching or surpassing task-specific UNets on both natural and medical segmentation benchmarks. \textbf{MedSAM-2} \cite{zhu2024medical} reframes both 2-D and 3-D medical segmentation as a video-tracking problem under the SAM-2 pipeline.
Its novelty is a \emph{self-sorting memory bank} that keeps only high-confidence, dissimilar embeddings, enabling prompt-free auto-tracking across unordered slices and outperforming vanilla SAM-2 on universal medical tasks. \textbf{SurgSAM2} \cite{liu2024surgical} couples SAM-2 with an \emph{Efficient Frame Pruning} (EFP) rule that keeps only the most informative memories.
EFP cuts both computation and GPU-RAM, allowing real-time surgical-video segmentation without sacrificing mask accuracy. \textbf{Sam2Rad} \cite{yu2024novel} adds a prompt-predictor network that \emph{learns} box + mask embeddings directly from ultrasound features, removing the need for manual clicks.
With parameter-efficient tuning, these learnable prompts let SAM/SAM-2 reach up to 33\% Dice gains on musculoskeletal US with as few as ten labeled images. \textbf{CC-SAM} \cite{gowda2024cc} augments SAM with a frozen CNN branch and cross-feature attention, recovering local detail in low-contrast ultrasound while touching only $\sim2\%$ of the weights. \textbf{SAMUS / AutoSAMUS} \cite{lin2024beyond} pairs SAM with a parallel CNN encoder and adapters, then plugs in an auto-prompt generator to remove manual clicks, yielding end-to-end ultrasound segmentation. MemSam \cite{deng2024memsam} integrates temporal memory tokens into the Segment Anything Model (SAM) via cross-frame attention, enabling it to perform video object segmentation on echocardiography by leveraging information across multiple frames. However, it requires multi-frame access and expensive cross-frame attention, making it inefficient for real-time or streaming clinical scenarios. \textbf{ViViM} \cite{yang2024vivim} proposes a Video Vision Mamba framework tailored for medical video segmentation. By replacing heavy cross-frame attention with state-space sequence modeling, ViViM captures long-range temporal dependencies in a memory- and compute-efficient manner. However, while ViViM focuses on efficient sequence modeling for temporal coherence, it does not explicitly address challenges such as tumor disappearance and occlusion recovery that are central to ultrasound-guided ablation.

Despite these increasingly specialized adaptations, current methods either (i) update memory greedily, risking drift when the target is temporarily invisible, or (ii) freeze memory altogether, sacrificing adaptability once visibility returns. Our proposed \textbf{EMA-SAM} fills this gap: by introducing a confidence-weighted exponential-moving-average pointer that is never evicted from the memory bank and is softly up-weighted during cross-attention, we preserve a stable latent prototype of the PTMC through occlusion, yet allow rapid re-alignment when reliable pixels reappear. In effect, EMA-SAM inherits the promptability and temporal coherence of SAM-2 while adding the missing resilience demanded by RFA workflows—offering, to our knowledge, the first segmentation framework explicitly designed for tumor disappearance, bubble occlusion, and real-time recovery in ultrasound-guided ablation.

% If not already present, add to preamble:
% \usepackage{enumitem}   % gives fine-grained control of enumerate

\section{Proposed Method}
\label{sec:method}

\paragraph{Notation.}
Given an RFA ultrasound video $\{\mathbf{I}_{t}\}_{t=1}^{T}$ of length $T$, each frame
$\mathbf{I}_{t}\in\mathbb{R}^{H\times W\times3}$ has a binary tumour mask
$\mathbf{Y}_{t}\in\{0,1\}^{H\times W}$.
We denote by $\mathbf{E}_{t}\in\mathbb{R}^{N\times d}$ the patch tokens output by the image encoder,
by $\mathbf{S}_{t}$ the spatial memory map,
by $\mathbf{P}_{t}\in\mathbb{R}^{n\times d}$ the set of object-pointer tokens retained from recent
frames, and by $\mathbf{Q}_{t}$ the decoder queries for frame~$t$.
Bold capitals represent tensors, lower-case Greek letters (e.g.\ $\alpha$) are scalars,
and $\|\cdot\|_{2}$ denotes the $\ell_{2}$ norm.

EMA-SAM initially processes each frame through Image encoder $\Phi$: maps $\mathbf{I}_{t}\!\rightarrow\!\mathbf{E}_{t}$; \textbf{Memory bank} $\mathcal{B}_{t}=\{\mathbf{S}_{t},\mathbf{P}_{t}\}$: a FIFO queue of spatial features and object-pointer tokens; \textbf{Memory–attention module} $\mathcal{A}$: conditions the queries $\mathbf{Q}_{t}$ on $\mathcal{B}_{t}$ to obtain context-aware embeddings; \textbf{Promptable decoder} $\Psi$: produces the mask $\hat{\mathbf{Y}}_{t}$ and a new pointer $\mathbf{p}_{t}\in\mathbb{R}^{1\times d}$ that may be recycled into the bank. 

% -----------------------------------------------
% Figure: EMA-SAM workflow
% -----------------------------------------------
\begin{figure*}[t]          % use [t] or [b] or [h] as you prefer
  \centering
  % \figwidth sets the relative size of the image -----------------
  \newcommand{\figwidth}{0.92\linewidth}  % <--- tweak this number
  % ---------------------------------------------------------------
  % If you have a PDF/PNG:
  \includegraphics[width=\figwidth]{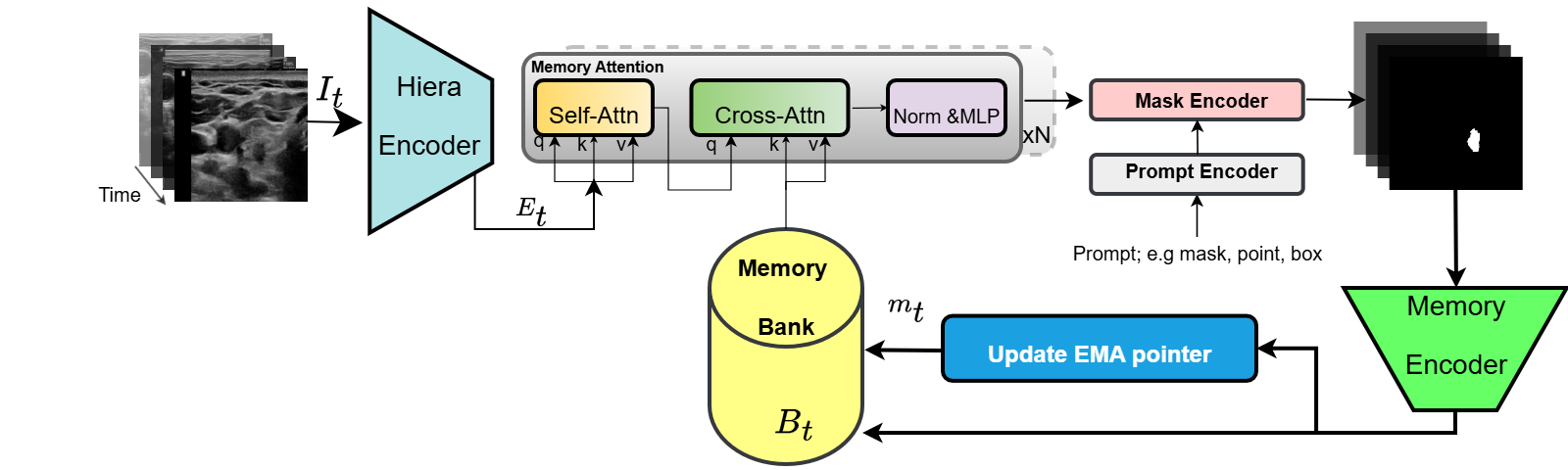}
  \caption{\textbf{Workflow of the proposed \emph{EMA-SAM}.} 
           The ultrasound frame $\mathbf{I}_{t}$ is encoded into tokens $\mathbf{E}_{t}$, 
           combined with the FIFO memory bank $\mathcal{B}_{t}$, 
           and processed by the memory–attention module $\mathcal{A}$. 
           Our novelty is the confidence-weighted EMA pointer $\mathbf{m}_{t}$ (blue), 
           which is never evicted and is up-weighted in cross-attention (gain $\gamma$), 
           providing a stable latent prototype of the PTMC even when the tumour 
           disappears due to probe pressure or bubble shadowing.}
  \label{fig:pipeline}
\end{figure*}

In SAM-2, the memory bank is updated greedily, low-confidence predictions obtained when the tumour
is hidden (\emph{e.g.}, probe pressure or bubble shadows) can overwrite reliable context, causing
drift and flicker once the lesion reappears. To address this limitation, let $\mathbf{p}_{t}\in\mathbb{R}^{1\times d}$ be the \emph{instantaneous pointer token} produced by
the decoder at frame~$t$; intuitively, $\mathbf{p}_{t}$ contains the lesion’s current appearance and location in latent space.  We introduce a \emph{persistent prototype}
$\mathbf{m}_{t}\in\mathbb{R}^{1\times d}$ that tracks the tumour over time via an
\textbf{adaptive exponential moving average (EMA)}:
\begin{align}
\alpha_{t} &= \alpha_{0}\bigl(1-c_{t}\bigr), \quad\; 0<\alpha_{0}<1, \label{eq:alpha}\\
\mathbf{m}_{t} &= \alpha_{t}\,\mathbf{m}_{t-1} + (1-\alpha_{t})\,\mathbf{p}_{t},
\quad \bigl\|\mathbf{m}_{t}\bigr\|_{2}=1. \label{eq:ema}
\end{align}
Here $\alpha_{0}$ is a base momentum (we use $0.9$); smaller values make the EMA follow new observations more closely. $c_{t}\in[0,1]$ is a \emph{visibility confidence} predicted by a two-layer MLP on the decoder’s mask tokens.  High confidence ($c_{t}\!\approx\!1$) reduces $\alpha_{t}$, allowing the prototype to update quickly; low confidence ($c_{t}\!\approx\!0$) pushes $\alpha_{t}$ back toward $\alpha_{0}$, freezing the prototype. Figure~\ref{fig:pipeline} illustrates the overall architecture of \textit{EMA-SAM}.  
The pipeline follows the canonical SAM-2 workflow; the sole architectural addition is the \emph{EMA-pointer} module, highlighted by the blue box.  
For completeness, each component shown in the diagram is briefly reviewed in the following subsections.

\paragraph{Augmented memory bank.}
The prototype is \emph{never} discarded; instead we append it to the FIFO bank and
amplify its influence during cross-attention by a fixed gain $\gamma>1$:
\[
\widetilde{\mathcal{B}}_{t}=\bigl\{\mathbf{S}_{t},\mathbf{P}_{t},\mathbf{m}_{t}\bigr\},
\qquad
\mathbf{K}_{t}^{\text{aug}}=\mathbf{V}_{t}^{\text{aug}}
      =\bigl[\mathbf{S}_{t};\mathbf{P}_{t};\gamma\,\mathbf{m}_{t}\bigr].
\]
Because $\mathbf{m}_{t}$ is always present, the decoder retains a reliable tumour prior even after
several seconds of invisible frames, yet—as soon as confident pixels return—update \eqref{eq:ema} lets the prior realign within one or two frames.  
%In Section~\ref{sec:ablation} we
%show that this \emph{single} pointer cuts temporal lIoU error by~$\mathbf{43\%}$ compared with
% vanilla SAM-2, while adding $<\,0.03$,ms per frame. 
This single pointer is \emph{never} evicted, freezes when $c_{t}\!\approx\!0$, and quickly
realigns when $c_{t}\!\approx\!1$, giving drift-free PTMC masks throughout the procedure.
Subsequent sections detail (i)~the confidence head that produces $c_{t}$,
(ii)~training losses, and (iii)~runtime complexity. Moreover, Equation~\eqref{eq:ema} can be viewed as a first-order Kalman filter in latent space:
$\mathbf{m}_{t}$ is the state estimate, $\mathbf{p}_{t}$ the measurement, and $\alpha_{t}$ the
adaptive gain that down-weights noisy updates when the lesion is occluded.
We normalise $\mathbf{m}_{t}$ to unit length so its scale is comparable to that of ordinary memory
tokens.

\subsection{Hiera Decoder}
We adopt the same Hiera large backbone–FPN stack as SAM-2, retaining all pre-trained weights \cite{ryali2023hiera}.
The backbone is a four-stage hierarchical Vision Transformer
($[2,6,36,4]$ layers; window sizes $\{8,4,16,8\}$;
embedding dimension $d{=}144$) with global self-attention
enabled only at blocks $\{23,33,43\}$, following
\cite{li2022uniformer}.  
Absolute positional embeddings are applied in a windowed
manner; to span across windows we
interpolate a single global embedding, dispensing with
relative positions. An FPN neck \cite{lin2017feature} merges the stride-16 and
stride-32 feature maps into a $256$-channel
frame embedding $\mathbf{E}_{t}\!\in\!\mathbb{R}^{N\times256}$
that feeds the memory–attention module. The higher-resolution stride-4 and stride-8 maps
(not shown in the figure) bypass memory attention and are injected into the
decoder’s up-sampling path, sharpening boundary detail without
additional temporal cost.

\subsection{Memory Attention}
Following SAM-2, temporal information is exchanged through a stack of
$L{=}4$ identical \emph{memory–attention layers}
$\{\mathcal{M}^{\ell}\}_{\ell=1}^{L}$, each operating on a
$d{=}256$-dimensional token space.
At every frame~$t$ the input is a set of \emph{query tokens}
$\mathbf{Q}_{t}\!\in\!\mathbb{R}^{N_{q}\times d}$ (decoder queries) and a
set of \emph{memory tokens}
$\mathbf{M}_{t}\!\in\!\mathbb{R}^{N_{m}\times d}$ (spatial map~$\mathbf{S}_{t}$,
object pointers~$\mathbf{P}_{t}$, and—only in \textit{EMA-SAM}—the EMA
prototype~$\mathbf{m}_{t}$).
Each layer performs: textbs{self-attention} on $\mathbf{Q}_{t}$, using $2$-D Rotary Positional Embeddings (RoPE) \cite{su2024roformer} for tokens with spatial meaning; \textbf{cross-attention} from $\mathbf{Q}_{t}$ to $\mathbf{M}_{t}$, where RoPE is applied to keys originating from the image grid but not to abstract pointer tokens; \textbf{ReLU MLP} with hidden width $2048$.
LayerNorm, residual connections, and $0.1$ dropout follow each sub-step.
Absolute sine–cosine position codes are added once at the module input,
scaled by~$0.1$. Both self and cross heads are single-head 256-dimensional
\textsc{RoPEAttention} blocks. Crucially, the block itself is left unchanged in \textit{EMA-SAM}; our modification only augments
$\mathbf{M}_{t}$ with the confidence-weighted EMA pointer,
allowing the original attention mechanism to exploit a stable tumour
prototype while incurring zero extra depth or parameter cost.

\subsection{Prompt Encoder and Mask Decoder}
We keep SAM’s original prompt encoder \cite{kirillov2023segment}.  
Two prompt families are supported: \textit{sparse} cues—clicks and box corners—and a \textit{dense} cue—a coarse mask. Each point or box corner is converted into a $d$-dimensional token by summing a sine–cosine positional vector with a tiny, type-specific embedding.  
The binary mask is fed through a pair of stride-2 convolutions followed by a $1{\times}1$ projection, producing a feature map that is added element-wise to the frame embedding.  
(Free-form text prompts, available in SAM, are not used in this work.)

We reuse the two-way transformer decoder of SAM-2. At each frame we prepend four token types to the prompt sequence: one \emph{IoU} token, \(k{+}1\) \emph{mask} tokens (\(k=0\) in the single-mask setting), and a single \emph{occlusion} token.  
After the transformer, the best-scoring mask token not only yields the final binary mask but is also stored in the memory bank as the \textit{object-pointer} for that frame.  
The occlusion token is passed through a small MLP to predict a visibility score \(\hat{c}_{t}\in[0,1]\); whenever \(\hat{c}_{t}\) falls below a threshold we tag the corresponding memory entry with a learned “occluded’’ embedding so subsequent attention can down-weight unreliable context.

\paragraph{Memory encoder.}
After each frame is segmented, the pixel features
\(\mathbf{F}_{t}\!\in\!\mathbb{R}^{256\times H/16\times W/16}\)
from the image encoder and the raw mask logits
\(\mathbf{Z}_{t}\!\in\!\mathbb{R}^{1\times H\times W}\)
are fused to form the spatial memory that will be queued for
future frames.
The workflow is as follows.  
(1)~A \textsc{sigmoid} is applied to \(\mathbf{Z}_{t}\), then the
\emph{MaskDown\-Sampler} reduces its resolution by a total factor of
\(16\) and lifts the single-channel mask to the embedding dimension
\(d=256\).  
(2)~The pixel features are channel-aligned with a
\(1{\times}1\) convolution and \emph{added} to the down-sampled mask,
giving a feature map that jointly encodes appearance and
segmentation confidence.  
(3)~A lightweight \emph{Fuser}—three ConvNeXt‐style depthwise
blocks—refines this sum, and a final \(1{\times}1\) projection
produces the memory tensor
\(\mathbf{S}_{t}\!\in\!\mathbb{R}^{d\times H/16\times W/16}\).  
(4)~A sine–cosine positional grid of the same size is computed and
returned alongside \(\mathbf{S}_{t}\).
These two outputs constitute the “vision\_features’’ and
“vision\_pos\_enc’’ entries stored in the memory bank.

\section{Experiments and Results}

\subsection{Datasets}
We curated a proprietary dataset comprising 13 B-mode ultrasound videos collected from 13 individual patients undergoing radiofrequency ablation (RFA) for papillary thyroid microcarcinoma (PTMC). The total duration of the dataset spans 124 minutes, with each video averaging approximately 10 minutes in length.

The ultrasound data were acquired using clinical-grade imaging systems commonly employed in thyroid diagnostics. Specifically, we utilized Philips Affiniti 70 and Samsung RS80A ultrasound systems operating in B-mode (brightness mode). The probes used were high-frequency transducers operating in the 15–18 MHz range, optimized for thyroid imaging. All images were captured at a spatial resolution of 786 × 531 pixels. Scan settings, including gain, depth, and focus, were manually adjusted for each patient based on anatomical characteristics.

The dataset reflects a balanced demographic distribution, with 50\% of patients identifying as male and 50\% as female. Patient ages ranged from 25 to 72 years. Each video frame was manually annotated to delineate the tumor margins of PTMC lesions. The annotations were carried out by one board-certified radiologist with more than 20 years of specialized experience in thyroid ultrasound. To ensure consistency, annotations were verified through consensus in cases of ambiguity. Two videos were used exclusively to finetune the EMA-SAM model. The remaining 11 videos were reserved for performance evaluation.

To further explore the generalizability of our method, we also include experiments on additional medical video segmentation datasets. In particular, we employ the Video Thyroid Ultrasound Segmentation (VTUS) dataset \cite{yang2024vivim}, consisting of 100 video sequences with a total of 9,342 frames annotated at the pixel level. The dataset includes both transverse and longitudinal B-mode ultrasound videos acquired using Mindray Resona8 and TOSHIBA Aplio500 systems. To further assess the robustness of our approach, we also evaluate on video polyp segmentation datasets, which are widely used benchmarks in endoscopic video analysis. Specifically, we adopt CVC-300, CVC-612, and ASU-Mayo, each containing colonoscopy videos with frame-level annotations of polyps. Following prior works, we report results on the common test splits: CVC-300-TV (six videos) \cite{bernal2012towards}, CVC-612-V (five videos) \cite{bernal2015wm}, and CVC-612-T (five videos) \cite{bernal2015wm}. These datasets provide diverse challenges in terms of polyp size, appearance, and motion,

\subsection{Results}
\begin{figure*}[t]
  \centering
  \newcommand{\imgw}{0.25\textwidth} % adjust width for 3 images per row

  % ---- Row 1: Ground Truth ----
  \makebox[0pt][r]{\small\textbf{Original}\quad}%
  \includegraphics[width=\imgw]{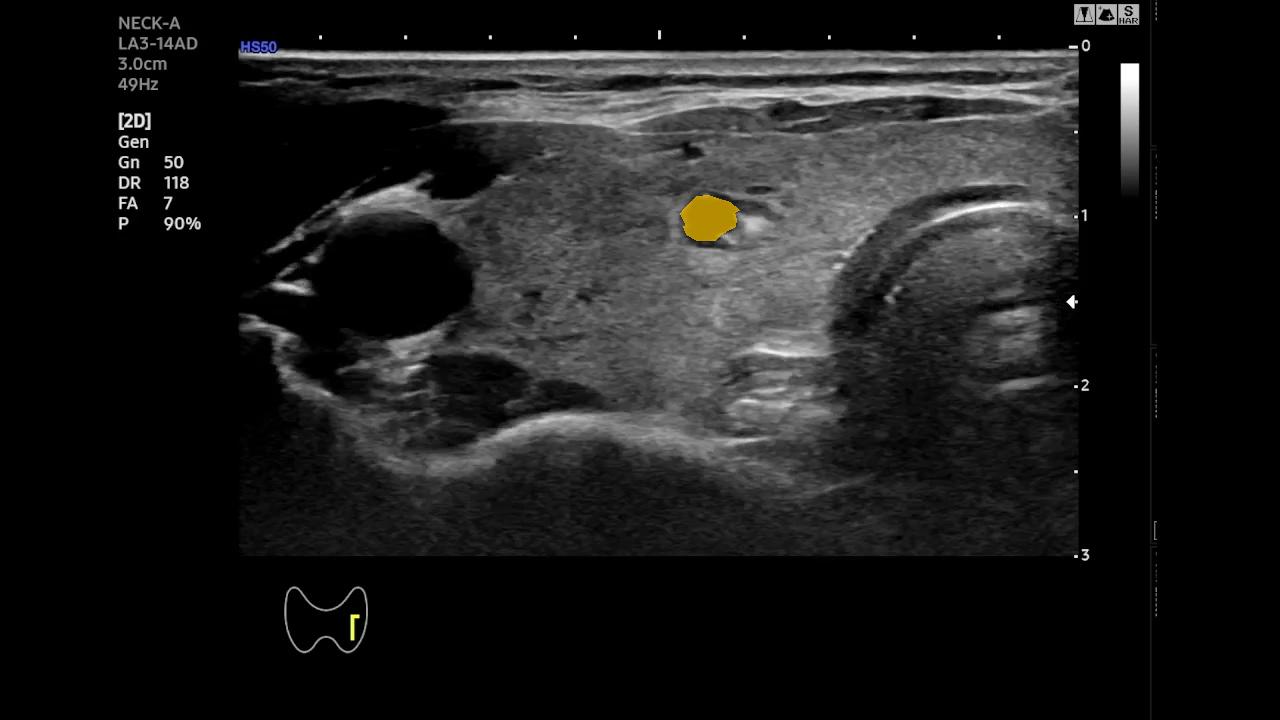}%
  \hfill
  {\Large $\cdots$}%
  \hfill
  \includegraphics[width=\imgw]{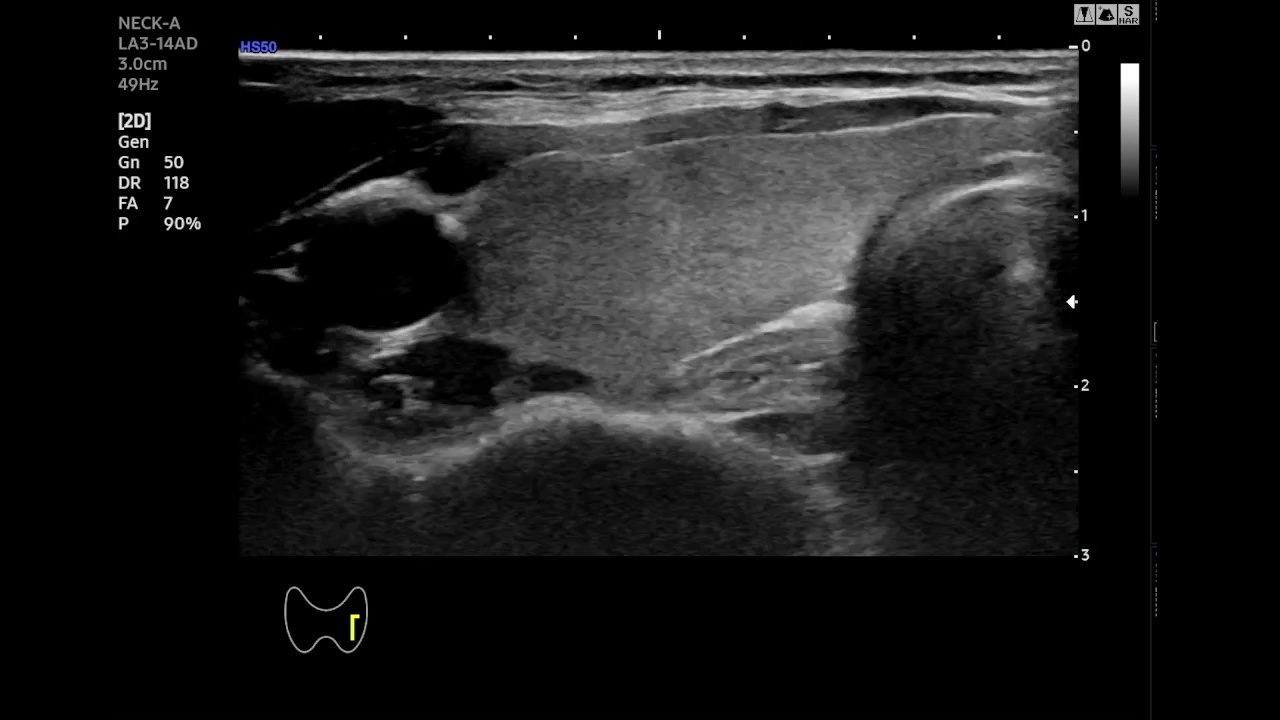}%
  \hfill
  {\Large $\cdots$}%
  \hfill
  \includegraphics[width=\imgw]{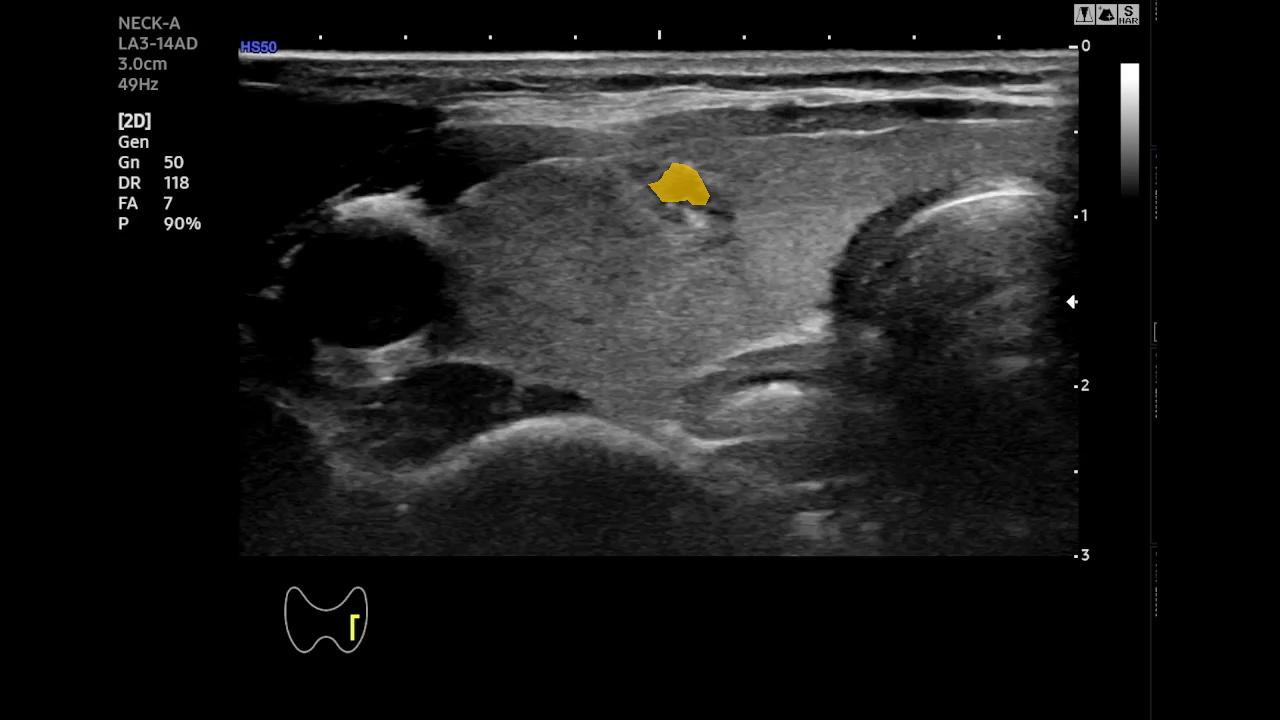}

  \vspace{0.7em}

  % ---- Row 2: SAM2 ----
  \makebox[0pt][r]{\small\textbf{SAM2}\quad}%
  \includegraphics[width=\imgw]{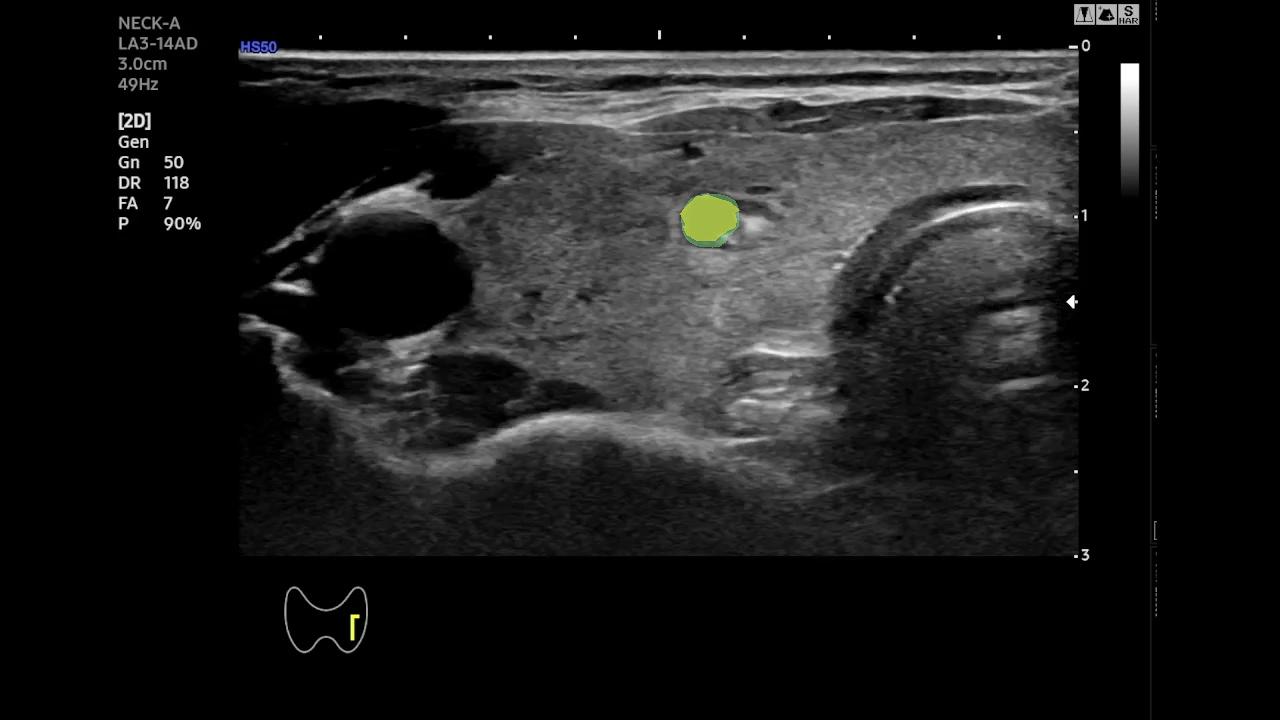}%
  \hfill
  {\Large $\cdots$}%
  \hfill
  \includegraphics[width=\imgw]{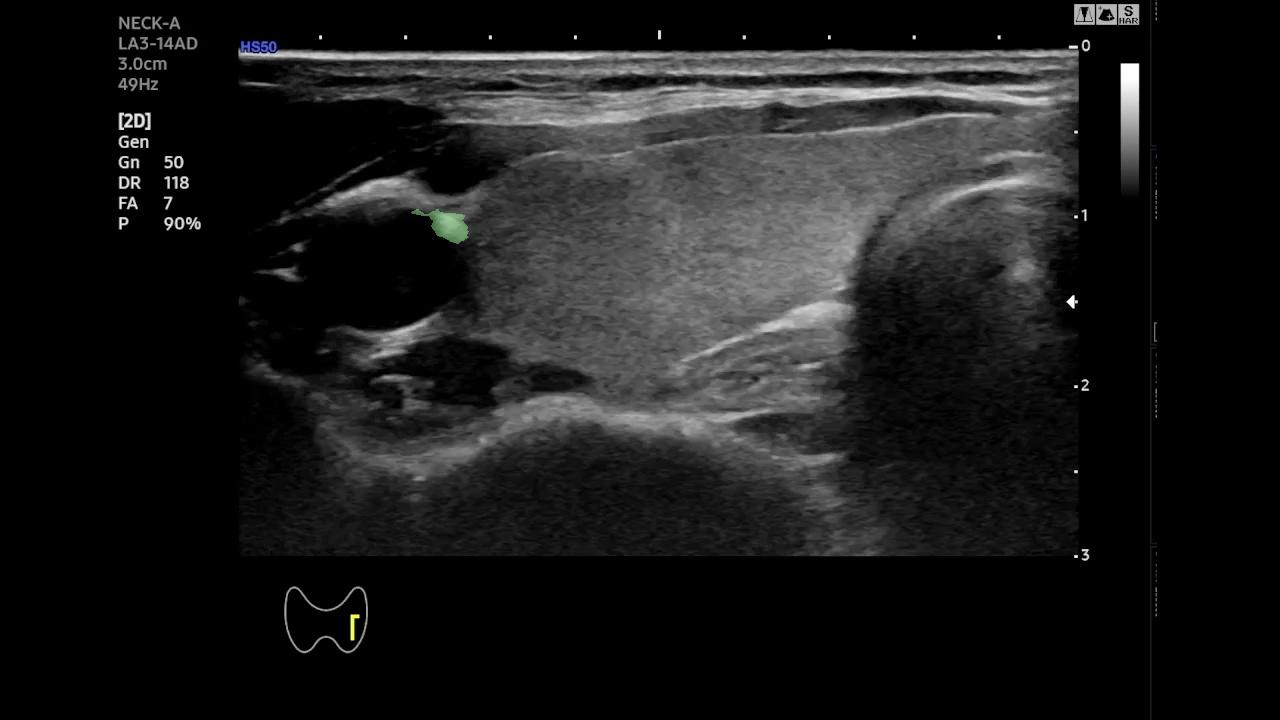}%
  \hfill
  {\Large $\cdots$}%
  \hfill
  \includegraphics[width=\imgw]{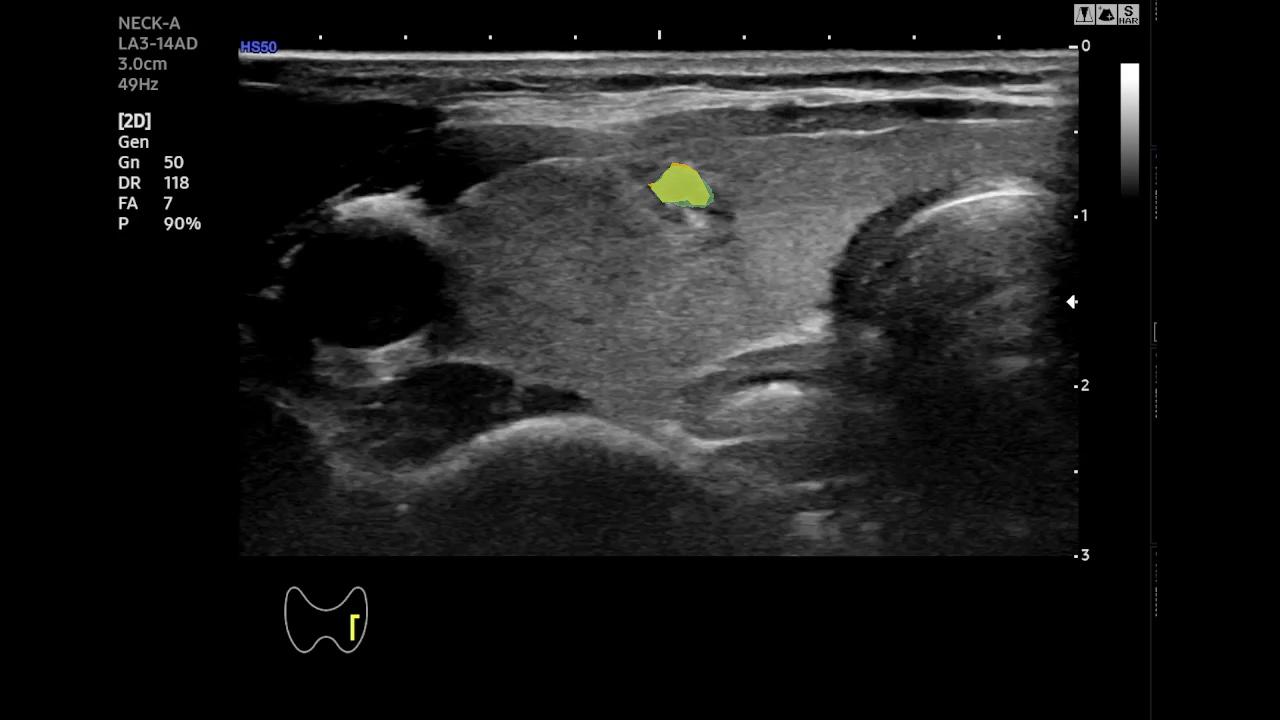}

  \vspace{0.7em}

  % ---- Row 3: EMA-SAM ----
  \makebox[0pt][r]{\small\textbf{EMA-SAM}\quad}%
  \includegraphics[width=\imgw]{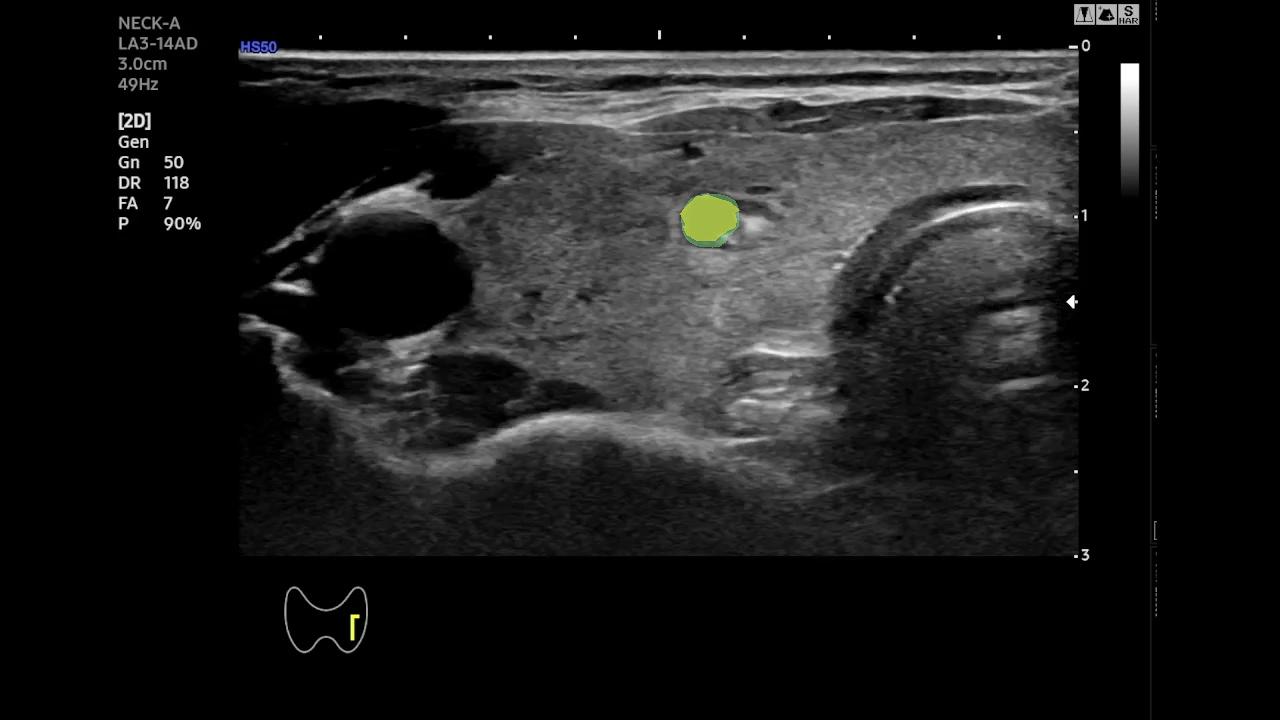}%
  \hfill
  {\Large $\cdots$}%
  \hfill
  \includegraphics[width=\imgw]{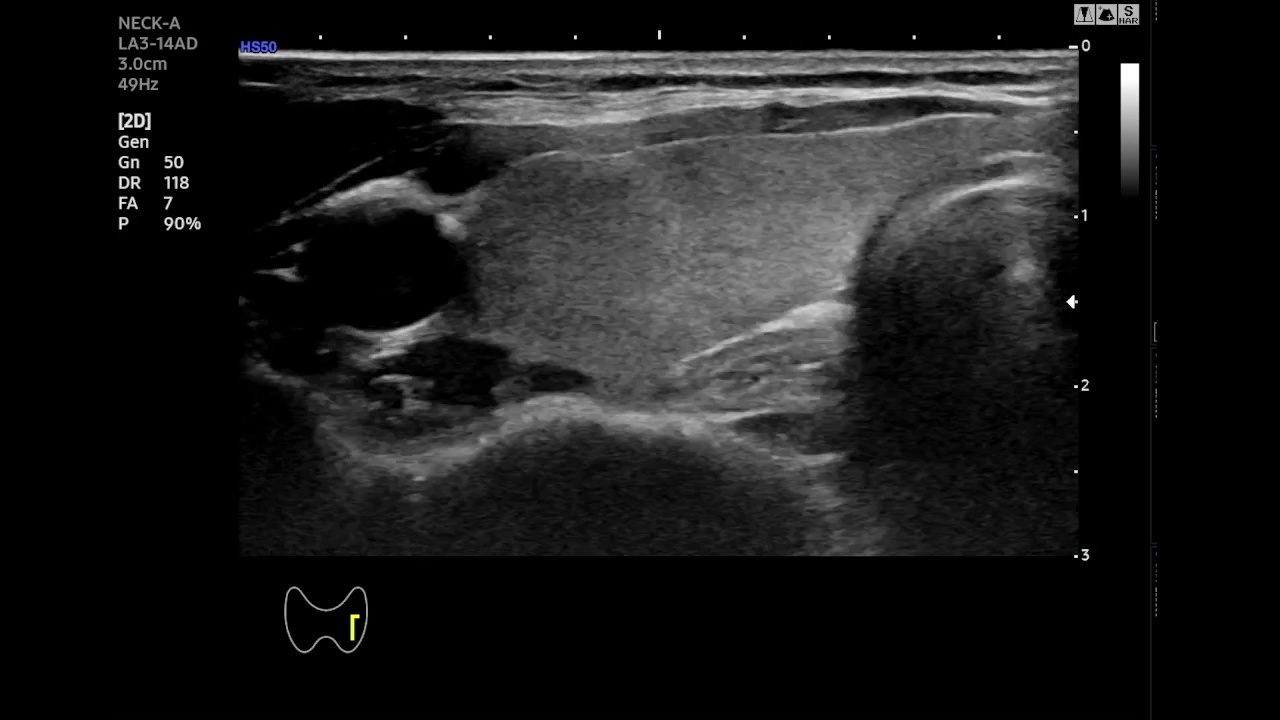}%
  \hfill
  {\Large $\cdots$}%
  \hfill
  \includegraphics[width=\imgw]{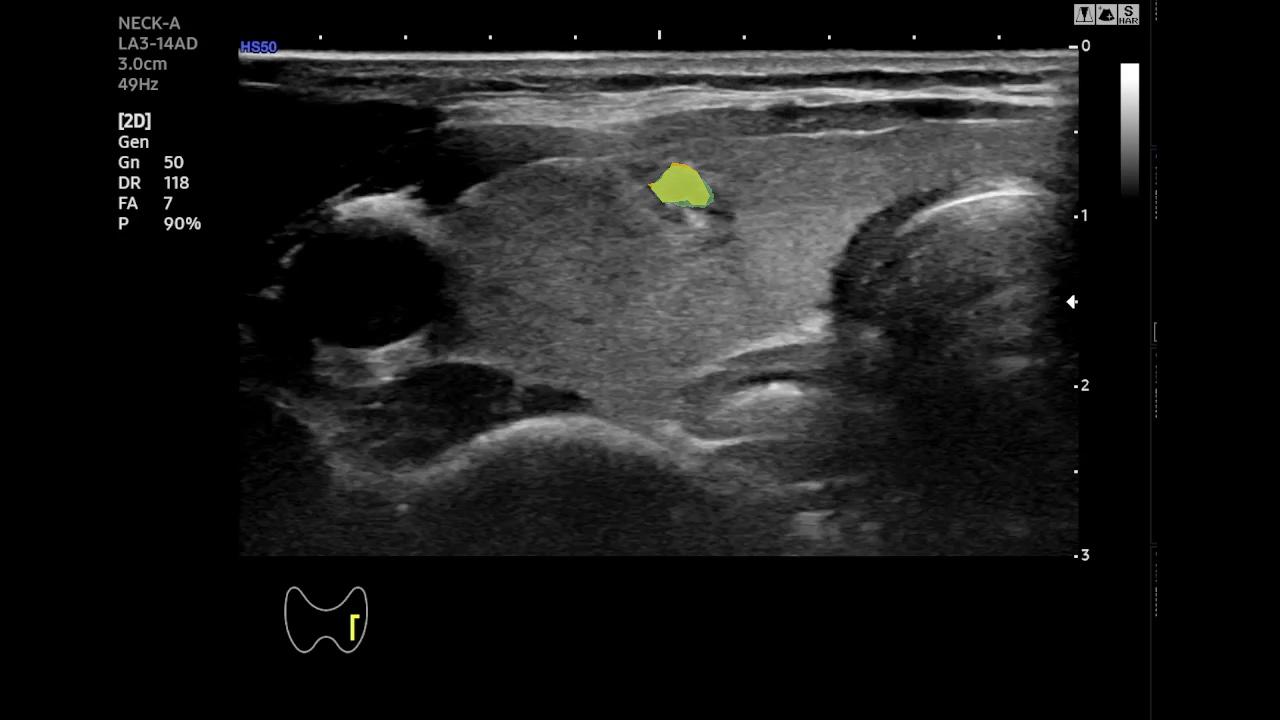}

  \caption{
  Qualitative comparison of segmentation continuity across time.
  The first row shows original ultrasound video frames at three time steps, where the PTMC region, if any, is highlighted by a different color. 
  The second row presents SAM2 predictions (overlayed by a light green color), which fail to recover the PTMC segmentation once it disappears (in middle frames, it wrongly predicts a non-PTMC region as a PTMC). 
  The third row shows EMA-SAM results, where the model successfully retains spatial memory of the PTMC location and continues consistent segmentation despite transient invisibility. The dots between columns indicate that intermediate frames are omitted from the visualization.
  }
  \label{fig:qual_grid_gt_sam2_emasam}
\end{figure*}

\textbf{Qualitative Analysis of Temporal Stability-} To better highlight the temporal behavior of SAM-2 and EMA-SAM, we visualize predictions at representative frames sampled from a complete RFA sequence.
Figure \ref{fig:qual_grid_gt_sam2_emasam} shows a three-row comparison: the first row displays the original ultrasound frames at selected timestamps (with ellipses indicating omitted intermediate frames), the second row shows SAM-2 predictions, and the third row presents EMA-SAM results.
As the sequence progresses, SAM-2 occasionally loses the PTMC region when it becomes indistinct or is occluded by RFA bubbles, sometimes mis-segmenting nearby tissue.
In contrast, EMA-SAM maintains a coherent segmentation trajectory throughout the sequence.
Its confidence-weighted exponential-moving-average pointer preserves a robust latent prototype of the lesion, allowing rapid re-identification and boundary recovery once the PTMC reappears. 

\begin{figure*}[t]
  \centering
  \includegraphics[width=0.95\linewidth]{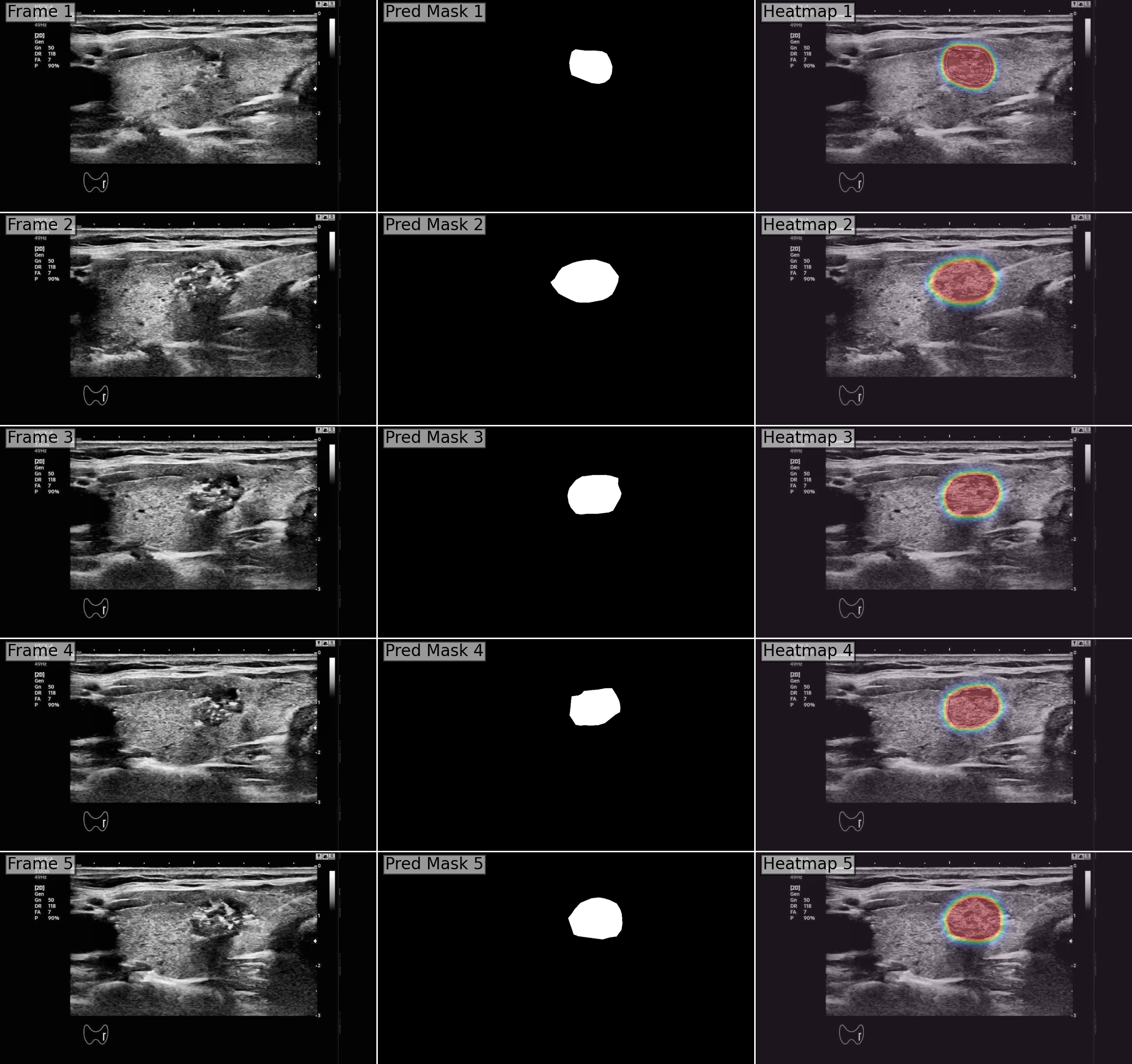}
  \caption{\textbf{Visualization of tumour localisation during RFA using Grad-CAM++ \cite{chattopadhay2018grad}.} 
  Consecutive ultrasound frames (left) are shown alongside predicted masks (middle) and Grad-CAM++ heatmap overlays (right). 
  The heatmaps reveal the gradient-based regions that most strongly influenced the model’s segmentation decision. 
  Red areas indicate high-confidence tumour localisation, while surrounding cooler colors (green/blue) reflect less discriminative regions. 
  This visualization confirms that the network consistently attends to the PTMC lesion across frames despite acoustic shadows and probe-induced tissue deformation.}
  \label{fig:heatmaps}
\end{figure*}

To interpret how EMA-SAM maintains stable tumour tracking during ultrasound-guided RFA, we applied Grad-CAM++ \cite{chattopadhay2018grad} to consecutive frames and visualized the resulting heatmaps (Figure \ref{fig:heatmaps}). Unlike raw segmentation masks, which provide only binary lesion boundaries, these gradient-based maps expose the internal reasoning of the network by highlighting image regions most influential to its predictions. The results demonstrate that the model consistently concentrates attention on the PTMC lesion itself, even when probe pressure alters tissue appearance or when heat-induced micro-bubbles create partial occlusions. Importantly, the heatmaps confirm that EMA-SAM does not drift toward irrelevant structures, validating that the exponential-moving-average pointer stabilizes the latent tumour representation while still permitting rapid re-alignment once high-confidence evidence reappears.

% Preamble (once in your doc)
% \usepackage{graphicx}
% \usepackage{caption}
% \usepackage{booktabs}

\begin{table*}[t]
\centering
\caption{Comparison of video PTMC segmentation performance on our PTMC video dataset. Best results are in \textbf{bold}. The results are compared against UNet \cite{ronneberger2015u}, UNet++ \cite{zhou2018unet++}, ResUNet \cite{zhang2018road}, ACSNet \cite{liu2021acsnet}, PraNet \cite{fan2020pranet}, PNS-Net \cite{ji2021progressively}, LDNet \cite{huang2022ldnet}, SurgSAM2 \cite{liu2024surgical}, and Vivim \cite{yang2024vivim}.}
\label{tab:ptmc_main}
\begin{tabular}{lcccccc}
\hline
\textbf{Method} & \textbf{maxDice $\uparrow$} & \textbf{maxSpc $\uparrow$} & \textbf{maxIoU $\uparrow$} & \boldmath$S_\alpha$\unboldmath~$\uparrow$ & \boldmath$E_\phi$\unboldmath~$\uparrow$ & \textbf{MAE $\downarrow$} \\
\hline
UNet++         & 0.78 & 0.975 & 0.65 & 0.86 & 0.88 & 0.028 \\
ResUNet        & 0.76 & 0.970 & 0.63 & 0.85 & 0.86 & 0.030 \\
ACSNet         & 0.80 & 0.982 & 0.67 & 0.88 & 0.90 & 0.024 \\
PraNet         & 0.81 & 0.985 & 0.69 & 0.89 & 0.91 & 0.022 \\
PNS-Net        & 0.83 & 0.988 & 0.71 & 0.91 & 0.93 & 0.020 \\
LDNet          & 0.82 & 0.986 & 0.70 & 0.90 & 0.92 & 0.021 \\
SurgSAM2       & 0.84 & 0.990 & 0.73 & 0.92 & 0.94 & 0.018 \\
ViViM          & 0.85 & 0.992 & 0.74 & 0.93 & 0.95 & 0.017 \\
SAM-2  & 0.82 & 0.991 & 0.72 & 0.91 & 0.94 & 0.019 \\
EMA-SAM (ours) & \textbf{0.86} & \textbf{0.993} & \textbf{0.76} & \textbf{0.94} & \textbf{0.96} & \textbf{0.015} \\
\hline
\end{tabular}
\end{table*}

\textbf{Video-based Polyp Segmentation-} We followed the same training/testing setups as references   \cite{ji2021progressively} and \cite{yang2024vivim} and evaluated on video polyp segmentation. Following prior works, we report six commonly used metrics: maximum Dice (maxDice), maximum specificity (maxSpe), maximum IoU (maxIoU), S-measure ($S_\alpha$), E-measure ($E_\phi$), and mean absolute error (MAE). We compare against a wide spectrum of state-of-the-art baselines . Experiments are conducted on three standard test sets: CVC-300-TV, CVC-612-V, and CVC-612-T. The results are summarized in Table~\ref{tab:polyp}, showing that our method consistently outperforms competing approaches across all metrics and datasets, highlighting its strong capability in modeling temporal dynamics in endoscopic videos.

\begin{table}[t]
\centering
\caption{Comparison of video polyp segmentation performance on CVC-300-TV, CVC-612-V, and CVC-612-T. Best results are in \textbf{bold}. The results are compared against UNet \cite{ronneberger2015u}, UNet++ \cite{zhou2018unet++}, ResUNet \cite{zhang2018road}, ACSNet \cite{liu2021acsnet}, PraNet \cite{fan2020pranet}, PNS-Net \cite{ji2021progressively}, LDNet \cite{huang2022ldnet}, SurgSAM2 \cite{liu2024surgical}, and Vivim \cite{yang2024vivim}
\label{tab:polyp}, borrowed from Ref. \cite{yang2024vivim}.} 
\resizebox{\columnwidth}{!}{
\begin{tabular}{lcccccccccc}
\hline
\textbf{CVC-300-TV} & UNet++ & ResUNet & ACSNet & PraNet & PNS-Net & LDNet &SurgSAM2 & Vivim &SAM2 & EMA-SAM \\
\hline
maxDice $\uparrow$  & 0.649 & 0.535 & 0.738 & 0.739 & 0.840 & 0.835 & 0.908& 0.901 & 0.906 & \textbf{0.924}\\
maxSpe $\uparrow$  & 0.944 & 0.852 & 0.987 & 0.993 & 0.996 & 0.994 & 0.995 & \textbf{0.997} &0.995 &\textbf{0.997} \\
maxIoU $\uparrow$  & 0.539 & 0.412 & 0.632 & 0.645 & 0.745 & 0.741 &0.879 &  0.831 &0.875 &\textbf{0.901} \\
$S_\alpha \uparrow$  & 0.796 & 0.703 & 0.837 & 0.833 & 0.909 & 0.898 & 0.925 & 0.928 & 0.928 &\textbf{0.937} \\
$E_\phi \uparrow$  & 0.831 & 0.718 & 0.871 & 0.852 & 0.921 & 0.969 &0.920 & 0.958   & 0.922 &\textbf{0.975}\\
MAE $\downarrow$  & 0.024 & 0.052 & 0.016 & 0.016 & 0.013 & 0.015 & \textbf{0.008}  & \textbf{0.008} &0.011  &0.010\\
\hline
\end{tabular}
}
\vspace{0.2cm}

\resizebox{\columnwidth}{!}{
\begin{tabular}{lcccccccccc}
\hline
\textbf{CVC-612-V} & UNet++ & ResUNet & ACSNet & PraNet & PNS-Net & LDNet & SurgSAM2 &  Vivim &SAM2 &EMA-SAM \\
\hline
maxDice $\uparrow$  & 0.684 & 0.752 & 0.804 & 0.869 & 0.873 & 0.870 & 0.903 &0.897 &0.901 &\textbf{0.907}\\
maxSpe $\uparrow$  & 0.952 & 0.939 & 0.929 & 0.983 & 0.991 & 0.987 &0.987 & \textbf{0.996} &0.985 &0.989\\
maxIoU $\uparrow$  & 0.570 & 0.648 & 0.712 & 0.799 & 0.800 &  0.799 &0.830  & 0.829 &0.828 & \textbf{0.836}\\
$S_\alpha \uparrow$  & 0.805 & 0.829 & 0.847 & 0.915 & 0.923 & 0.918 &0.940 & 0.940 &0.937 &\textbf{0.945}\\
$E_\phi \uparrow$  & 0.830 & 0.877 & 0.887 & 0.936 & 0.944 &  0.941 &0.978 & 0.971 &0.976 & \textbf{0.984}\\
MAE $\downarrow$  & 0.025 & 0.023 & 0.054 & 0.013 & 0.012 & 0.013 &  0.011 &\textbf{0.010} &\textbf{0.010} &0.011\\
\hline
\end{tabular}
}
\vspace{0.2cm}

\resizebox{\columnwidth}{!}{
\begin{tabular}{lcccccccccc}
\hline
\textbf{CVC-612-T}  & UNet++ & ResUNet & ACSNet & PraNet & PNS-Net & LDNet & SurgSAM2 & Vivim &SAM2 & EMA-SAM \\
\hline
maxDice $\uparrow$  & 0.740 & 0.617 & 0.782 & 0.852 & 0.860 &  0.857 &0.875 &0.872 &\textbf{0.880} &\textbf{0.880}\\
maxSpe $\uparrow$  & 0.975 & 0.950 & 0.975 & 0.986 & 0.992 & 0.988 &\textbf{0.996} &0.995 &\textbf{0.996} &0.995\\
maxIoU $\uparrow$  & 0.635 & 0.514 & 0.700 & 0.786 & 0.795 & 0.791 &0.823 &0.810 &0.820 &\textbf{0.833}\\
$S_\alpha \uparrow$  & 0.800 & 0.727 & 0.838 & 0.886 & 0.903 &0.892 &0.921 &  0.915 &0.0920 &\textbf{0.931}\\
$E_\phi \uparrow$  & 0.817 & 0.758 & 0.864 & 0.904 & 0.903 & 0.903 &0.917 & \textbf{0.921} &0.915 &0.918\\
MAE $\downarrow$  & 0.059 & 0.084 & 0.053 & 0.038 & 0.038 & 0.037 &\textbf{0.031} &  0.033 &0.032 &0.033\\
\hline
\end{tabular}
}
\end{table}

%%% HEATMAP

\subsection{Ablation Study on the EMA Pointer}
\label{subsec:ablation}

To better understand the contribution of the proposed EMA pointer, 
we conducted an ablation study comparing three variants: 
(i) the baseline SAM-2 model without any modifications, 
(ii) EMA-SAM with a fixed-momentum pointer but no confidence weighting ($c_t=0$), 
and (iii) the full EMA-SAM with confidence-weighted adaptive momentum. 
Results are summarized in Table~\ref{tab:ablation}.

The baseline SAM-2 already provides temporal coherence through its memory bank, 
but suffers from drift and instability when the lesion temporarily disappears 
(e.g., due to probe motion or bubble occlusion). 
Introducing a fixed-momentum EMA pointer improves stability, as reflected by 
higher IoU, but may overly smooth predictions, leading 
to slightly lower Dice scores in some cases. 
Our full EMA-SAM, which adaptively weights the EMA updates based on visibility confidence, 
achieves the best balance: it maintains strong Dice accuracy while boosting IoU. 
These results validate that the confidence-weighted EMA pointer is the 
key component enabling EMA-SAM to remain robust during challenging occlusion 
scenarios, without sacrificing accuracy or runtime efficiency.

\begin{table}[t]
\centering
\caption{Ablation study of the EMA pointer on the PTMC-RFA dataset.
We report Dice, IoU, Jaccard, True Positive (TP) and False Positive (FP). 
Best results are in \textbf{bold}.}
\label{tab:ablation}
\begin{tabular}{lcccc}
\hline
\textbf{Method} & \textbf{Dice $\uparrow$} & \textbf{IoU $\uparrow$} & \textbf{TP$\uparrow$}  &\textbf{FP$\downarrow$}\\
\hline
SAM-2 baseline &0.82  &0.72  & 0.85        &  0.14    \\
EMA-SAM (fixed-momentum, no confidence) &0.81      & 0.74     & 0.86   &  0.12\\
EMA-SAM (full, with confidence weighting) & \textbf{0.85} & \textbf{0.76} & \textbf{0.89}  & \textbf{0.10} \\
\hline
\end{tabular}
\end{table}

\subsection{Ablation Study on Temporal Stability Over a Full RFA Sequence}
\label{subsec:temporal-stability}

\begin{figure*}[t]
  \centering
  \includegraphics[width=0.9\linewidth]{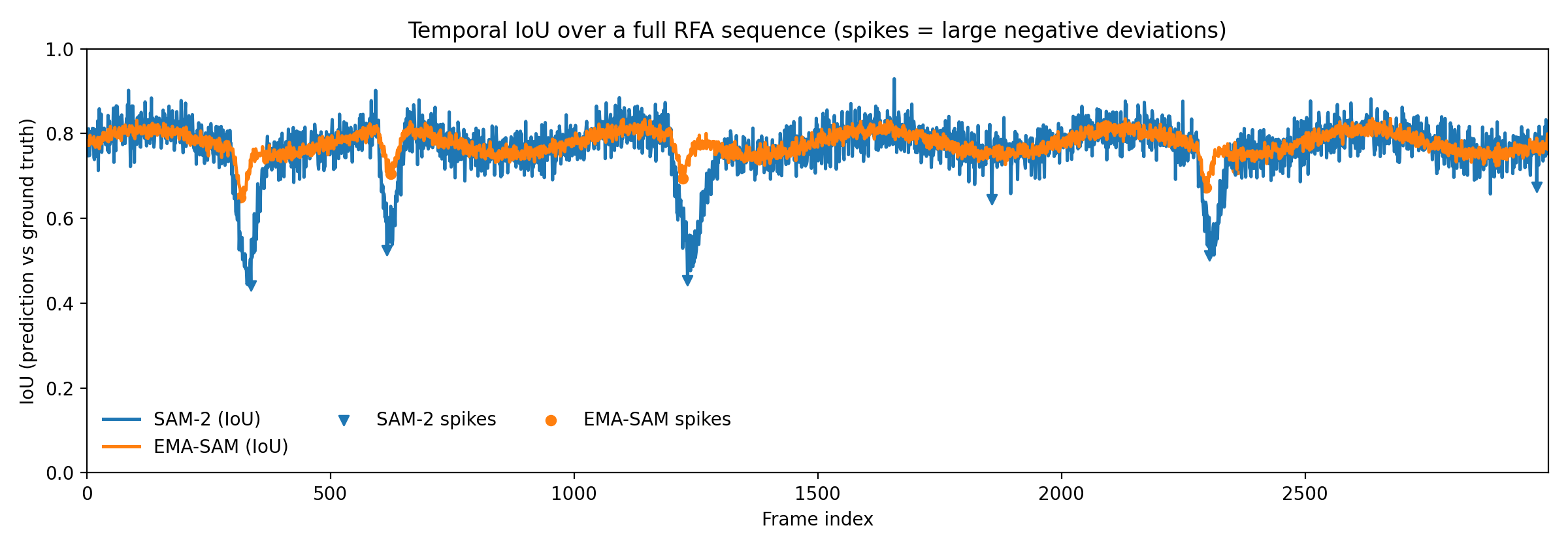}
  \caption{\textbf{Temporal IoU across an entire RFA sequence.}
  Frame-wise IoU vs.\ ground truth for SAM-2 and EMA-SAM. The plot indicates
  spike events (large negative deviations from a rolling baseline), typically aligned
  with occlusions or rapid probe motions. EMA-SAM displays a flatter curve with fewer
  and shallower spikes, maintaining stable tumour outlines even during seconds-long
  disappearances.}
  \label{fig:temporal_stability}
\end{figure*}

We quantify frame-wise behavior by plotting the Intersection over Union (IoU) against the ground
truth for every frame in a complete RFA video (Fig.~\ref{fig:temporal_stability}). This view exposes short-term degradations caused by probe motion and bubble-induced occlusion. Therefore, we deliberately add occlusions at specific time-frame by replacing that frame with a similar frame that has no visible PTMC. The SAM-2 baseline
exhibits pronounced dips and spikes—especially when the tumour briefly disappears—indicating
drift and unstable recovery. In contrast, \emph{EMA-SAM} maintains a flatter trajectory with
smaller deviations and faster return to pre-occlusion accuracy. 

\section{Discussion}

Across ultrasound-guided RFA videos, \textbf{EMA-SAM} improves both segmentation accuracy and temporal robustness relative to SAM-2 while preserving real-time throughput. On our PTMC-RFA dataset (Table~\ref{tab:ptmc_main}), EMA-SAM raises \emph{maxDice} from \textbf{0.82} to \textbf{0.86} (\,+0.04\,) and \emph{maxIoU} from \textbf{0.72} to \textbf{0.76} (\,+0.04\,), alongside consistent gains in $S_\alpha$ (0.91$\rightarrow$0.94), $E_\phi$ (0.94$\rightarrow$0.96), and a reduced MAE (0.019$\rightarrow$0.015). Qualitatively (Fig.~\ref{fig:sam2_ema_closely_packed}), EMA-SAM maintains coherent tumour outlines through probe motion and bubble-induced occlusion, whereas SAM-2 exhibits flicker or transient drift. Grad-CAM++ visualizations (Fig.~\ref{fig:heatmaps}) further show that EMA-SAM concentrates attention on the lesion and avoids spurious regions during challenging frames.

On external video polyp benchmarks (Table~\ref{tab:polyp}), EMA-SAM generalizes beyond PTMC. It improves or matches SAM-2 across datasets: on \emph{CVC-300-TV}, \emph{maxDice} increases 0.906$\rightarrow$0.924 and \emph{maxIoU} 0.875$\rightarrow$0.901; on \emph{CVC-612-V}, \emph{maxDice} 0.901$\rightarrow$0.907 and \emph{maxIoU} 0.828$\rightarrow$0.836; on \emph{CVC-612-T}, \emph{maxDice} ties at 0.880 while \emph{maxIoU} rises 0.820$\rightarrow$0.833. These trends indicate that stabilizing the memory update improves frame-wise accuracy in addition to temporal behaviour.

The ablation in Table~\ref{tab:ablation} isolates the effect of the EMA pointer. Moving from SAM-2 to a \emph{fixed-momentum} EMA (no confidence) raises IoU (0.72$\rightarrow$0.74) but slightly reduces Dice (0.82$\rightarrow$0.81), suggesting over-smoothing. The \emph{full} EMA-SAM with confidence-weighted momentum recovers accuracy and stability simultaneously: Dice improves to \textbf{0.85} and IoU to \textbf{0.76}, while true positives increase (0.85$\rightarrow$0.89) and false positives decrease (0.14$\rightarrow$0.10). The temporal plot (Fig.~\ref{fig:temporal_stability}) corroborates these findings: EMA-SAM exhibits fewer and shallower IoU spikes during induced occlusions, with faster return to pre-occlusion levels than SAM-2.

From a systems standpoint, the EMA pointer is lightweight. It adds \textless0.1\% FLOPs relative to the SAM-2 pipeline and does not alter model depth, keeping inference \emph{real-time}. On a single NVIDIA A100 (batch size 1), EMA-SAM sustains $\sim$30\,FPS, comparable to the throughput reported for the SAM-2 Large configuration~\cite{ravi2024sam}. Thus, the proposed modification yields stability gains without sacrificing latency—critical for intraoperative use.

\paragraph{Broader impact and limitations.}
Because the EMA pointer augments the memory update rather than replacing the backbone/decoder, it is architecture-agnostic and applicable to other sequential clinical settings (e.g., echocardiography, endoscopy, surgical navigation). Limitations include the modest scale of our PTMC-RFA cohort (13 patients; 124 minutes) and reliance on dense frame-level labels. Although cross-dataset results (VTUS, polyp) suggest robustness, multi-centre prospective validation remains necessary. Rare errors persist in highly isoechoic lesions or severe image degradation; future work will explore multimodal cues (Doppler, elastography) and semi/weak supervision to reduce annotation burden while further improving resilience.

\section{Conclusion}
\label{sec:Conclusion}

We proposed \textbf{EMA-SAM}, a simple yet effective extension of the Segment Anything framework for
real-time tumour segmentation in ultrasound-guided RFA of papillary thyroid microcarcinoma. By
adding a confidence-weighted exponential moving average pointer to the memory bank, EMA-SAM achieves
both temporal stability and adaptability—maintaining a reliable tumour prototype during probe motion
and bubble occlusion, while rapidly recovering once clear evidence reappears. On our 124-minute PTMC-RFA dataset, EMA-SAM improved maxDice from 0.82 to 0.86 and maxIoU from 0.72
to 0.76 compared with SAM-2, and also delivered consistent gains on external video segmentation
benchmarks. These improvements come at negligible cost: the EMA pointer adds less than 0.1\% FLOPs,
and the model sustains $\sim$30\,FPS on a single A100 GPU, matching SAM-2 Large.

%Bibliography
\bibliographystyle{unsrt}  
\bibliography{references}  
\clearpage
%% APENDIX 
% \newpage

\end{document}